# Hybrid Reasoning Network for Video-based Commonsense Captioning

Weijiang Yu[1], Jian Liang[2], Lei Ji[3], Lu Li[4], Yuejian Fang[2]†, Nong Xiao[1]†, Nan Duan[3]
[1]Sun Yat-sen University, [2]Peking University, [3]Microsoft Research Asia, [4]Zhejiang University
weijiangyu8@gmail.com,j.liang@stu.pku.edu.cn,leiji@microsoft.com,lu.lee@zju.edu.cn
fangyj@ss.pku.edu.cn,xiaon6@mail.sysu.edu.cn,nanduan@microsoft.com

## ABSTRACT

The task of video-based commonsense captioning aims to generate event-wise captions and meanwhile provide multiple commonsense descriptions (e.g., attribute, effect and intention) about the underlying event in the video. Prior works explore the commonsense captions by using separate networks for different commonsense types, which is time-consuming and lacks mining the interaction of different commonsense. In this paper, we propose a Hybrid Reasoning Network (HybridNet) to endow the neural networks with the capability of semantic-level reasoning and word-level reasoning. Firstly, we develop multi-commonsense learning for semantic-level reasoning by jointly training different commonsense types in a unified network, which encourages the interaction between the clues of multiple commonsense descriptions, event-wise captions and videos. Then, there are two steps to achieve the word-level reasoning: (1) a memory module records the history predicted sequence from the previous generation processes; (2) a memory-routed multi-head attention (MMHA) module updates the word-level attention maps by incorporating the history information from the memory module into the transformer decoder for word-level reasoning. Moreover, the multimodal features are used to make full use of diverse knowledge for commonsense reasoning. Experiments and abundant analysis on the large-scale Video-to-Commonsense benchmark show that our HybridNet achieves state-of-the-art performance compared with other methods.

## 1 INTRODUCTION

Recently, research on video-based commonsense captioning [11] has been gaining increasing attention, as it provides a deeper understanding of the video and language and thus facilitates various visual reasoning tasks ranging from fundamental scene understanding [8, 19, 42] to high-level visual-linguistic reasoning tasks [13, 18, 44, 46]. The video-based commonsense captioning task aims to generate captions and three types of commonsense descriptions (intention, effect, attribute) simultaneously given the input video. A show case is presented in Figure 1 (a).

The video-based commonsense captioning is a frontier research topic and the current best performer [11] executes separate networks to learn different types of commonsense separately. It is time-consuming and counterintuitive. When humans infer a specific event, each commonsense is not identified individually. We often consider a global perspective to reason the commonsense semantic coherence by interacting with the clues from various commonsense types. For example, humans can correctly reason about the intention in Figure 1 (b) not only benefit from the video input and event-wise caption but also with the help of accurate predictions of attribute and effect. Such high-level semantic interaction can facilitate multiple commonsense reasoning.

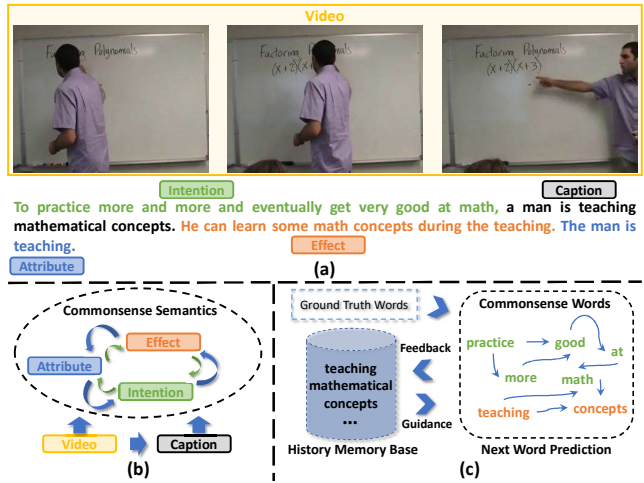

Figure 1: (a) The example of the video-based commonsense captioning task. It requires the network to not only generate event-wise captions but also provide visually grounded commonsense descriptions about the underlying event in the video. The task seeks to describe the intentions of the agent, effects of the action, and the attributes of the agent's characteristic. (b) Semantic-level reasoning to interplay with different commonsense clues. (c) Word-level reasoning generates the next word, which is guided by the history sequences as well as using the ground truth words during training.

Besides, humans can remain a logic loop when performing commonsense inference. Taking the intention sentence as an example in Figure 1 (a), when humans see the history token sequence like "to practice more and more and eventually get very", we can easily infer the next word should be positive (e.g., "good"). The "good" can also be used to reason the accuracy of previously predicted words and then correct them for contextual logic consistency (i.e. logic loop). As we can see, such word-level reasoning in the description can benefit from each predicted word. However, current methods [11, 34, 45] follow the Transformer [35] training mechanism to predict the next word only with the ground truth words as context while generating the entire sequence from scratch at inference, which leads to a gap between the previously predicted words and the next word.

To address the above issues, we achieve the semantic-level reasoning in Figure 1 (b) to jointly learn the multiple commonsense types. Then we perform the word-level reasoning in Figure 1 (c) in the network for bridging the gap between the previously predicted words and the next word. In this paper, we propose a novel Hybrid Reasoning Network (HybridNet) in Figure 2 (a) to subtly integrate these two levels of reasoning into a unified framework.

For semantic-level reasoning, we propose multi-commonsense learning as shown in Figure 2 (a) to achieve semantic-level reasoning, in which multiple learning commonsense descriptions are solved at the same time while exploiting commonalities and differences across commonsense semantics. Multi-commonsense learning is a mechanism to inductive transfer that improves generalization by using the domain information contained in the training signals of related commonsense as an inductive bias. It achieves this by reasoning various types of commonsense descriptions in parallel while using a shared representation. What is learned for each commonsense semantic can help other commonsense semantics be learned better. Furthermore, we merge 3D motion features, 1D audio features and 2D appearance features together via the multimodal fusion for enforcing our model to learn to reason commonsense semantics from diverse information.

To utilize the previous context words for predicting the next word, we propose a memory module to record the history information from previous generation processes. Then we introduce a memory-routed multi-head attention (MMHA) to incorporate the history information with the ground truth words into the network for next word prediction during the training. Specifically, our MMHA infers the next word conditioned on history information and ground truth words by learning a merging attention map. In Figure 3 (b), the MMHA first predicts the conditional attention map routed by the previous states from the memory module. Then we design a triangle convolution to learn the conditional attention map and meanwhile prevent foreseeing subsequent positions. Next, we merge the conditional attention map with several attention maps, including the original attention map from the multi-head attention and the previous attention maps of the front blocks. The previous attention maps of different blocks are bridged via our proposed contextual residual connections to learn the long-range attention context. Finally, our MMHA updates the word feature (i.e. V) via a merged attention map to achieve word-level reasoning. In this paper, we utilize these two modules to achieve word-level reasoning. Their architectures can be seen in Figure 2 (b) and Figure 3 (b).

**Contributions. (1)** We propose a Hybrid Reasoning Network (HybridNet) to jointly perform semantic-level reasoning and word-level reasoning. **(2)** A multi-commonsense learning is proposed to achieve the semantic-level reasoning based on the video and caption inputs. **(3)** A memory-routed multi-head attention is introduced to cooperate with a novel memory module to execute the word-level reasoning. **(4)** Extensive experiments and abundant analysis have demonstrated the effectiveness and superiority of our proposed modules.

## 2 RELATED WORK

**Visual Comprehension.** More and more researchers pay more attention to the visual comprehension community by targeting visual question answering [1], visual dialogue [9], visual question generation [20], image captioning [38], visual commonsense reasoning [44], visual grounding [27], situation recognition [42] and video captioning [23]. Recently, the visual reasoning tends to cognition-level reasoning by incorporating the commonsense concepts [4, 5, 16, 31–33] from the natural language processing (NLP). For example, early works [26, 39] utilized the prior commonsense knowledge to assist in the prediction of action motivation. Zellers et al. [44] proposed a visual commonsense reasoning task to not only provide question answering but also predict the correct rationale behind the answer based on the question and image. Based on the cause-effect clues, Wang et al. [40] presented a visual commonsense R-CNN on object detection by mining the commonsense knowledge behind the object categories. Yu et al. [43] proposed a heterogeneous graph learning method to seamlessly integrate the intra-graph and inter-graph reasoning in order to bridge multi-modal domains for visual commonsense reasoning. For video-based commonsense captioning, Fang et al. [11] used individual transformers for different commonsense captioning, which lacks of the commonsense interaction. In this work, we propose a hybrid reasoning network to jointly learn multiple commonsense descriptions via semantic-level reasoning and word-level reasoning.

**Video Captioning.** Visual perception and language expression are two key capabilities of human intelligence, and video captioning is an insight example towards learning from humans to bridge vision and language. There are mainly two aspects for video captioning development: datasets and methods. For developing the video captioning, some early works proposed some specific-domain datasets like movie [28, 30] and cooking [10, 29], which is limited and small for deep learning. Some researchers tended to open-domain video captioning datasets such as MSVD [7], MSR-VTT [41] and TGIF [21]. Recently, video captioning tries to connect with the commonsense to explore the commonsense descriptions in the video, which comes up with a dataset named Video-to-Commonsense [11]. In the method sight, current deep-learning-based video captioning often performs sequence-to-sequence learning in an encoder-decoder paradigm. In between, an encoder equipped with powerful deep neural networks is exploited to learn video representation. A decoder of sentence generation is utilized to translate the learned representation into a sentence with more flexible structures. Venugopalan et al. [37] applied the sequence-to-sequence model into the video captioning by end-to-end learning way. To bridge the sentence semantics and visual content, an attention-based LSTM named aLSTMs [12] is proposed to better transfer videos to natural sentences by capturing salient structures of video. For dense video captioning, Zhou et al. [45] used an end-to-end transformer [35] architecture to jointly learning the video encoder, proposal decoder, and captioning decoder. Fang et al. [11] used an encoder-decoder way to individually model the specific commonsense captioning without using the commonsense correlations. In this paper, we propose to generate the commonsense descriptions in the video from semantic-level and word-level inference.

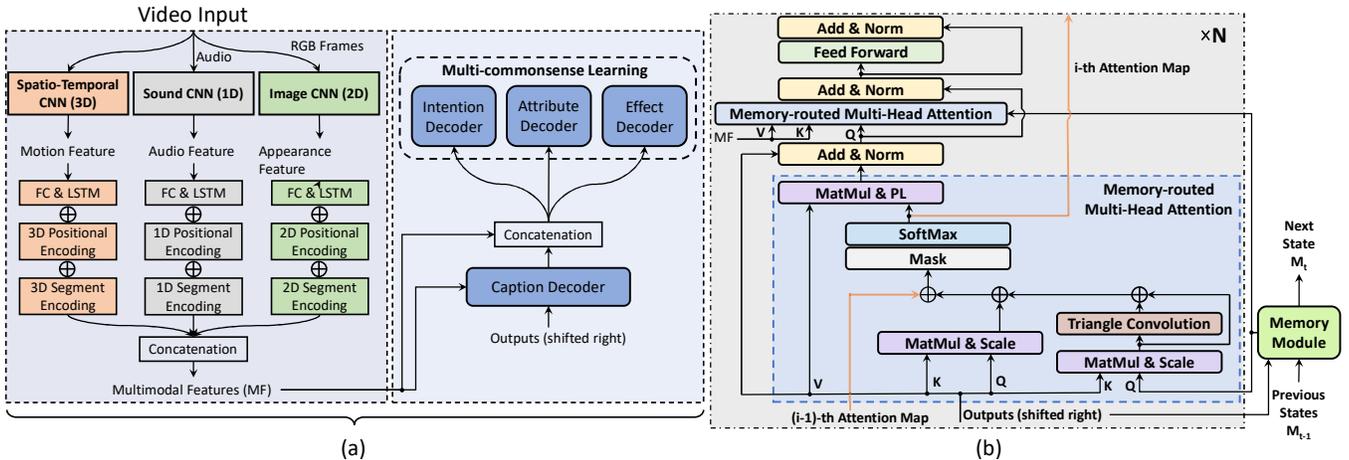

Figure 2: (a) Overview of our HybridNet. Taking the video as input, a multimodal fusion is shown to merge the motion feature, audio feature and appearance feature as multimodal features (MF). Then the MF is fed into the decoder stage for captioning, including the caption decoder and three commonsense decoders. We utilize multi-commonsense learning on three commonsense decoders. (b) The pipeline of the proposed memory-routed multi-head attention (MMHA) in each decoder block. The MMHA is guided by the memory module that is illustrated in Figure 3 (b). The "'MatMul & Scale' indicates the scaled dot-product operation and the "PL" means the projection layer. The orange lines with arrows denote the contextual residual connections to bridge contextual attention maps from different blocks. The $\bigoplus$ means the addition operation in this work.

## 3 HYBRID REASONING NETWORK

In this section, we explain the architecture and design of our proposed Hybrid Reasoning Network (HybridNet), which can appropriately interact with multiple commonsense semantics and preserve the connection between contextual predicted sequences. As shown in Figure 2 (a), our HybridNet is an encoder-decoder architecture, including a video encoder, a caption decoder and three commonsense decoders. Given a video input, there are three pre-trained models to extract multiple features including motion feature, audio feature and appearance feature. Then a multimodal fusion is utilized to merge the extracted features as multimodal features. The multimodal features are fed into the caption decoder to obtain caption encoding as well as predict event-wise captions. Then we concatenate the multimodal features with caption encoding as the input of commonsense decoder to generate the commonsense descriptions. Note that the multi-commonsense learning is applied on the commonsense decoder. And the decoder blocks are enhanced by our memory-routed multi-head attention (MMHA), memory module and contextual residual connections. Our contribution mainly focuses on the design of the multi-commonsense learning and innovative modules (e.g., MMHA and memory module), which are unveiled and discussed in detail in the following sub-sessions.

There are two settings for the two sub-tasks. **Completion task:** the ground truth caption and video are given to generate the commonsense descriptions. **Generation task:** given the video input, the event-wise captions should be predicted first, and then use both of them to predict the desired commonsense sentences.

### 3.1 Encoder

Given a video, we use the pre-trained models including ResNet152 [14], SoundNet [2] and I3D [6] to extract the appearance feature, audio feature and motion feature, respectively. Then we use a multimodal fusion to merge the three types of features. As shown in Figure 2 (a)(left), we use separate linear layers (FC) and LSTM [15] to individually encode the different features, and utilize the last hidden states of the LSTM as the final representations. Finally, the customized positional encoding and segment encoding are added to the final representations, which are concatenated together as the multimodal features. Taking the motion feature as an example

$$E^{3D} = SE^{3D} + PE^{3D} + \text{LSTM}(\text{FC}(V^{3D})), \quad (1)$$

where $E^{3D}$ is the encoded motion feature and $V^{3D}$ means motion feature. The $SE^{3D}$ and $PE^{3D}$ are 3D segment encoding and 3D positional encoding, respectively. Similarly, we can obtain the encoded audio feature $E^{1D}$ and encoded appearance feature $E^{2D}$. Then we concatenate them together to get the multimodal features (MF).

### 3.2 Decoder

In our decoder, we propose two main innovations: (1) multiple commonsense learning for semantic-level reasoning; (2) memory-routed multi-head attention cooperated with memory module for word-level reasoning. The first one is to improve the high-level inference ability from various commonsense semantics. The second one aims to mine the low-level reasoning from different words.

### 3.3 Semantic-level Reasoning

**Multi-commonsense Learning.** Multi-commonsense learning is a training paradigm in which machine learning models are trained with data from multiple types of commonsense descriptions simultaneously, using shared representations to learn the common ideas between a collection of related commonsense. These shared representations increase data efficiency and can potentially yield a faster learning speed for correlated descriptions. In this paper, we develop the transformer based language model [35] as three commonsense decoders for three particular commonsense domains. There are many different factors to consider when creating a shared architecture for multi-commonsense learning, such as the portion of the model's parameters that will be shared between commonsense, and how to parameterize and combine commonsense-specific and shared modules. In our HybridNet, we share the parameters of encoder and caption decoder for all commonsense decoders. During the training, the commonsense decoder takes the video encoding $\mathbf{v}$, caption encoding $\hat{\mathbf{s}}$ and ground truth of corresponding commonsense captions (e.g., $\mathbf{c_{att}}$, $\mathbf{c_{eff}}$, $\mathbf{c_{int}}$) as input to iteratively generate commonsense descriptions, which can be formulated as

$$\hat{\mathbf{c}}_{\mathbf{att}} = \mathbf{D}_{\text{ATT}}(\mathbf{v}, \hat{\mathbf{s}}, \mathbf{c_{att}}), \tag{2}$$

$$\hat{\mathbf{c}}_{\mathbf{eff}} = \mathbf{D}_{\text{EFF}}(\mathbf{v}, \hat{\mathbf{s}}, \mathbf{c_{eff}}), \tag{3}$$

$$\hat{\mathbf{c}}_{\mathbf{int}} = \mathbf{D}_{\text{INT}}(\mathbf{v}, \hat{\mathbf{s}}, \mathbf{c_{int}}), \tag{4}$$

where $\hat{\mathbf{c}}_{\mathbf{att}}$, $\hat{\mathbf{c}}_{\mathbf{eff}}$, $\hat{\mathbf{c}}_{\mathbf{int}}$ are the generated commonsense sequences decoded by the corresponding commonsense decoders ($\mathbf{D}_{\text{ATT}}$, $\mathbf{D}_{\text{EFF}}$, $\mathbf{D}_{\text{INT}}$). The loss function for the $\mathbf{D}_{\text{ATT}}$ can be formulated as

$$\mathcal{L}_{att} = \sum_{t=1}^{N_{att}} \log p(\mathbf{y}_t | \mathbf{y}_{t-1}, [\mathbf{v}, \hat{\mathbf{s}}]; \Theta_{\mathbf{att}}), \tag{5}$$

where $\mathbf{y}_t$ denotes the one-hot vector probability of each word at time $t$, $N_{att}$ denotes the length of the attribute. The attribute decoder parameters $\Theta_{\mathbf{att}}$ are trained to maximize the log-likelihood over the training set. Similarly, we can obtain the objection functions of effect decoder and intention decoder like $\mathcal{L}_{eff}$ and $\mathcal{L}_{int}$. The caption decoder can be optimized by

$$\mathcal{L}_{cap} = \sum_{t=1}^{N_{cap}} \log p(\mathbf{y}_t | \mathbf{y}_{t-1}, \mathbf{v}; \Theta_{\mathbf{cap}}). \tag{6}$$

Finally, we train our HybridNet by using $\mathcal{L} = \mathcal{L}_{cap} + \mathcal{L}_{att} + \mathcal{L}_{eff} + \mathcal{L}_{int}$ to jointly optimize our framework.

### 3.4 Word-level Reasoning

**Memory Module.** For any relevant videos, they may share similar patterns in their descriptions that can be used as good references for each other to help the generation process. Besides, the previous sequence can be recorded to guide the next word prediction for contextual consistency. To exploit such characteristics, we propose to use an extra component named memory module to enhance Transformer to learn from the previous word information and facilitate computing the interactions among previous information and the generation process.

As shown in Figure 3(b), our memory module uses a matrix $\mathbf{M}$ to transfer its states over generation steps, where the states record the important word information with each row (namely, memory slot) representing some word information.[1] During the generation, the matrix is updated step-by-step by incorporating the output from previous steps. Then, at time step $t$, the matrix from the previous step, $\mathbf{M}_{t-1}$, is functionalized as the query and its concatenations with the previous output serve as the key and value

---
[1]Note that the rows (memory slots) and word states do not follow one-to-one mapping, where the entire matrix serves as a whole unit to deliver the word information.

to feed into the multi-head attention module. Given $H$ heads used in Transformer, there are $H$ sets of queries, keys and values via three linear transformations, respectively. For each head, we obtain the query, key and value in the memory module through $\mathbf{Q} = \mathbf{M}_{t-1}\mathbf{W_q}$, $\mathbf{K} = [\mathbf{M_{t-1}}; \mathbf{y_{t-1}}]\mathbf{W_k}$ and $\mathbf{V} = [\mathbf{M_{t-1}}; \mathbf{y_{t-1}}]\mathbf{W_v}$, respectively, where $\mathbf{y}_{t-1}$ is the embedding of the last output (at step $t-1$); $[\mathbf{M_{t-1}}; \mathbf{y_{t-1}}]$ is the row-wise concatenation of $\mathbf{M}_{t-1}$ and $\mathbf{y}_{t-1}$. The $\mathbf{W_q}$, $\mathbf{W_k}$ and $\mathbf{W_v}$ are the trainable weights of linear transformation of the query, key and value, respectively. Multi-head attention is used to model $\mathbf{Q}$, $\mathbf{K}$ and $\mathbf{V}$ so as to depict relations of different patterns. As a result,

$$\mathbf{Z} = \text{softmax}(\mathbf{Q}\mathbf{K}^\top/\sqrt{d_k})\mathbf{V}, \tag{7}$$

where $d_k$ is the dimension of $\mathbf{K}$, and $\mathbf{Z}$ the output of the multi-head attention module. Consider that the memory module is performed in a recurrent manner along with the decoding process, it potentially suffers from gradient vanishing and exploding. Therefore, we introduce residual connections and a series of gate operations. The former is formulated as

$$\mathbf{M'}_t = f_{mlp}(\mathbf{Z} + \mathbf{M}_{t-1}) + \mathbf{Z} + \mathbf{M}_{t-1}, \tag{8}$$

where $f_{mlp}(\cdot)$ refers to the multi-layer perceptron (MLP). Moreover, we apply the forget and input gates to balance the inputs from $\mathbf{M}_{t-1}$ and $\mathbf{y}_{t-1}$, respectively. To ensure that $\mathbf{y}_{t-1}$ can be used for computation with $\mathbf{M}_{t-1}$, it is extended to a matrix $\mathbf{Y}_{t-1}$ by duplicating it to multiple rows. Therefore, the forget gate $\mathbf{G}_t^f$ and input gate $\mathbf{G}_t^i$ are formalized as

$$\mathbf{G}_t^f = \mathbf{Y}_{t-1}\mathbf{W}^f + \tanh(\mathbf{M}_{t-1})\mathbf{U}^f, \tag{9}$$

$$\mathbf{G}_t^i = \mathbf{Y}_{t-1}\mathbf{W}^i + \tanh(\mathbf{M}_{t-1})\mathbf{U}^i, \tag{10}$$

where $\mathbf{W}^f$ and $\mathbf{W}^i$ are trainable weights for $\mathbf{Y}_{t-1}$ in each gate; similarly, $\mathbf{U}^f$ and $\mathbf{U}^i$ are the trainable weights for $\mathbf{M}_{t-1}$ in each gate. The final output of the gate mechanism is formalized as

$$\mathbf{M}_t = \sigma(\mathbf{G}_t^f) \odot \mathbf{M}_{t-1} + \sigma(\mathbf{G}_t^i) \odot \tanh(\mathbf{M'}_t), \tag{11}$$

where $\odot$ refers to the Hadamard product and $\sigma$ is the sigmoid function. The $\mathbf{M}_t$ is the output of the entire memory module at step $t$, which is fed into the MMHA for routing the decoder.

**Memory-routed Multi-Head Attention (MMHA).** We design the MMHA in each decoder block to bridge the previous word states and the next predicting word. We think the previous words can cooperate with the multimodal features to better reason the generation processes. As shown in Figure 2 (b), given the input representation $\mathbf{X}$, we get the query $\mathbf{Q}$, key $\mathbf{K}$ and value $\mathbf{V}$ through three linear layers. Then a multi-head scale dot-product operation is utilized to generate the original attention map

$$\mathbf{A}_o = \text{Attention}(\mathbf{X}) = \mathbf{Q}\mathbf{K}^\top/\sqrt{d_k}, \tag{12}$$

where $\mathbf{A}_o$ denotes the original attention map and $d_k$ is the hidden dimension size of $\mathbf{K}$. Because of the guidance of the memory module, we can obtain the memory state $\mathbf{M}$ from the memory module that records the previous sequence state. The $\mathbf{M}$ is regarded as another $\mathbf{Q}$ in the MMHA. Then another multi-head scale dot-product operation is applied to getting the conditional attention map $\mathbf{A}_c$. Assume there are $K$ heads in each layer, then we get $K$ conditional attention maps. They construct a tensor $\mathbf{A}_c \in \mathbb{R}^{N \times N \times K}$ ($N$ is the sequence length), which can be viewed as a $N \times N$ image with $K$ channels. Taking this as input, we adopt one 2D-convolutional layer with 3 × 3 kernels to capture the evolution of attention patterns, as this inductive bias emphasizes local details and produces more precise attention maps by reasoning on previous ones. The output channel is also set to be $K$, so the attention maps of all heads can be generated jointly.

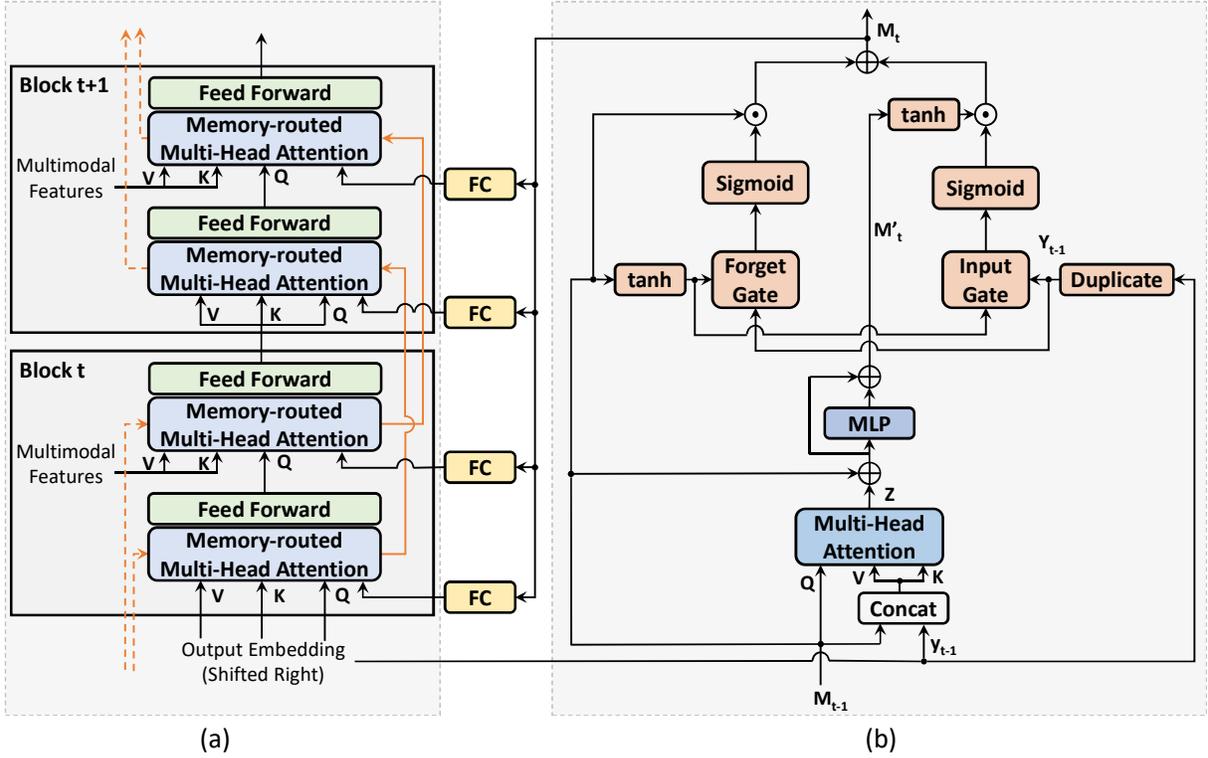

Figure 3: (a) Our residual connections that are marked as orange propagate the contextual information between different decoder blocks. Note that we omit additions and norm in the figure for brevity. (b) Our memory module records the information from previous generation processes. $\odot$ denotes the Hadamard product.

To prevent foreseeing subsequent positions, we improve the 2D-convolutional layer by proposing a triangle convolution. The triangle convolution can be implemented as follows: (1) executing standard 3 × 3 convolution with masks in the upper-right corner; (2) after convolution, shifting the entire attention matrix by one pixel to the bottom and one pixel to the right. We apply a ReLU activation after each 2D-convolution layer to provide non-linearity and sparsity. After the triangle convolution, the result attention map $\mathbf{A}_{triangle}$ is combined with the input conditional attention map $\mathbf{A}_c$, original attention map $\mathbf{A}_o$ and the attention maps from previous blocks $\mathbf{A}_{previous}$. Mathematically,

$$\begin{aligned} \mathbf{A}_{condition} &= \alpha \cdot \mathbf{A}_{triangle} + (1-\alpha) \cdot \mathbf{A}_c \quad, \\ \mathbf{A}_{guide} &= \beta \cdot \mathbf{A}_{condition} + (1-\beta) \cdot \mathbf{A}_o \quad, \\ \mathbf{A}_{merge} &= \gamma \cdot \mathbf{A}_{previous} + (1-\gamma) \cdot \mathbf{A}_{guide} \quad, \end{aligned} \quad (13)$$

where $\alpha, \beta, \gamma \in [0,1]$ are hyper-parameters for linear combination. In our experiments, the values of $\alpha$, $\beta$ and $\gamma$ are chosen empirically on the validation set for each task. Note that the $\mathbf{A}_{previous}$ is the residual attention map of previous blocks via contextual residual connections, which can be seen in Figure 3 (a). These contextual residual connections aim to bridge the gap of attention maps between different blocks for improving the quality of attention maps globally. Finally, the output of the MMHA can be obtained by

$$\mathbf{A}_{out} = \text{softmax}(\text{MASK}(\mathbf{A}_{merge})), \quad (14)$$
$$\mathbf{X}_{out} = \text{FC}(\mathbf{A}_{out}\mathbf{V}^\top), \quad (15)$$

where $\mathbf{A}_{out}$ is the attention map of current MMHA and the $\mathbf{X}_{out}$ means the output representation of the MMHA.

## 4 EXPERIMENTS
### 4.1 Datasets and Evaluation

We evaluate our proposed HybridNet and compare it with other state-of-the-art methods on Video-to-Commonsense (V2C) [11] benchmark, a representative large-scale video-based commonsense captioning dataset containing a total of 121,618 captions derived from 9,721 video scenes. The dataset is officially split into a training set consisting of 6819 videos with 85,100 captions, a test set containing 2903 videos with 36,518 captions. We follow this data partition in all experiments. We have done some statistics on the V2C and found that there are 5 candidate descriptions per video for intention, effect, and attribute respectively. And the number of candidate captions for each video is not fixed. For the training set, there are 766 samples contain 1~4 captions per video. The number of the video containing 5~14 captions is 3082. The test set also has the same challenge. Hence, the V2C is challenging because of the complex and diverse language, multiple scenes and hard inference types as mentioned in [11]. Followed other works [11, 12, 37, 45], for two sub-tasks, we measure the performance of our proposed method via Meteor [3], Rouge [22], CIDEr [36] and BLEU (n=1-4) [24].

Table 1: Evaluation of V2C completion task and generation task in terms of the attribute, effect and intention by using CIDER, BLEU, Rouge, and Meteor metrics. We use only BLEU-1 to evaluate the attribute generation on the completion task since the average length of the ground truth is just less than 2. "Attribute+C" means the attribute descriptions and the predicted event-wise captions on the generation task. The best performing results are marked in red.

|  | Relation | Model | CIDER | BLEU-1 | BLEU-2 | BLEU-3 | BLEU-4 | METEOR | ROUGE-L |
|---|---|---|---|---|---|---|---|---|---|
| Completion Task | Attribute | S2VT [37] | - | 35.9 | - | - | - | - | - |
| | | Attention-Enc-Dec [12] | - | 38.3 | - | - | - | - | - |
| | | Dense Captioner [45] | - | 46.0 | - | - | - | - | - |
| | | Video CMS Transformer [11] | - | 47.3 | - | - | - | - | - |
| | | Our HybridNet | - | **58.7**$^{+11.4}$ | - | - | - | - | - |
| | Effect | S2VT [37] | 28.3 | 24.9 | 18.6 | 16.2 | 14.3 | 15.4 | 22.1 |
| | | Attention-Enc-Dec [12] | 29.5 | 26.5 | 19.4 | 18.8 | 15.1 | 17.5 | 23.9 |
| | | Dense Captioner [45] | 36.9 | 33.7 | 24.8 | 21.0 | 20.2 | 20.0 | 29.9 |
| | | Video CMS Transformer [11] | 37.3 | 34.8 | 25.9 | 22.5 | 20.4 | 20.8 | 30.6 |
| | | Our HybridNet | **66.2**$^{+28.9}$ | **49.0**$^{+14.2}$ | **42.9**$^{+17.0}$ | **40.3**$^{+17.8}$ | **38.8**$^{+18.4}$ | **30.0**$^{+9.2}$ | **41.5**$^{+10.9}$ |
| | Intention | S2VT [37] | 51.8 | 48.4 | 39.9 | 34.3 | 26.4 | 23.3 | 44.3 |
| | | Attention-Enc-Dec [12] | 52.1 | 51.1 | 42.6 | 35.5 | 28.2 | 24.3 | 48.0 |
| | | Dense Captioner [45] | 60.3 | 59.3 | 47.0 | 37.3 | 31.5 | 28.0 | 53.1 |
| | | Video CMS Transformer [11] | 62.0 | 60.8 | 48.4 | 39.1 | 34.1 | 28.5 | 54.6 |
| | | Our HybridNet | **92.6**$^{+30.6}$ | **69.4**$^{+8.6}$ | **60.5**$^{+12.1}$ | **55.4**$^{+16.3}$ | **53.1**$^{+19.0}$ | **35.8**$^{+7.3}$ | **60.1**$^{+5.5}$ |
| Generation Task | Attribute+C | S2VT [37] | 38.5 | 69.1 | 53.6 | 42.0 | 32.3 | 23.9 | 59.1 |
| | | Attention-Enc-Dec [12] | 34.0 | 67.0 | 51.7 | 40.7 | 31.4 | 23.3 | 58.0 |
| | | Dense Captioner [45] | 36.8 | 68.4 | 52.1 | 39.8 | 30.0 | 24.1 | 57.7 |
| | | Video CMS Transformer [11] | 40.2 | 70.2 | 54.8 | 42.7 | 32.6 | 24.7 | 59.0 |
| | | Our HybridNet | **41.6**$^{+1.4}$ | **71.3**$^{+1.1}$ | **57.0**$^{+2.2}$ | **45.6**$^{+2.9}$ | **35.7**$^{+3.1}$ | **25.5**$^{+0.8}$ | **60.4**$^{+1.4}$ |
| | Effect+C | S2VT [37] | 29.9 | 69.8 | 54.1 | 39.9 | 29.1 | 21.9 | 55.3 |
| | | Attention-Enc-Dec [12] | 26.1 | 70.2 | 51.8 | 38.6 | 28.7 | 22.5 | 53.5 |
| | | Dense Captioner [45] | 30.6 | 72.1 | 54.5 | 42.3 | 33.2 | 25.2 | 56.0 |
| | | Video CMS Transformer [11] | 32.1 | 72.5 | 56.1 | 44.3 | 35.2 | 25.6 | 57.4 |
| | | Our HybridNet | **34.2**$^{+2.1}$ | **73.2**$^{+0.7}$ | **57.4**$^{+1.3}$ | **46.3**$^{+2.0}$ | **37.2**$^{+2.0}$ | **26.3**$^{+0.7}$ | **58.3**$^{+0.9}$ |
| | Intention+C | S2VT [37] | 35.4 | 71.3 | 53.9 | 41.3 | 31.2 | 21.6 | 58.6 |
| | | Attention-Enc-Dec [12] | 33.2 | 75.4 | 59.4 | 45.1 | 33.5 | 24.6 | 59.6 |
| | | Dense Captioner [45] | 37.0 | 76.1 | 60.2 | 46.7 | 35.9 | 26.5 | 60.9 |
| | | Video CMS Transformer [11] | 37.8 | 76.2 | 61.2 | 48.1 | 37.3 | 26.9 | 61.9 |
| | | Our HybridNet | **40.4**$^{+2.6}$ | **77.5**$^{+1.3}$ | **62.9**$^{+1.7}$ | **50.4**$^{+2.3}$ | **40.2**$^{+2.9}$ | **27.8**$^{+0.9}$ | **62.9**$^{+1.0}$ |

### 4.2 Implementation Details

We conduct all experiments by using a single NVIDIA 3090 card on a single server. We implement our proposed HybridNet and re-implement other state-of-the-art methods via PyTorch [25] and python3.8 to train and test. The Nvidia CUDA of 11.1 and cuDNN of 8.0 are utilized for acceleration. In our HybridNet, each decoder consists of 6 transformer blocks with 8 attention heads. Note that the traditional multi-head attention is replaced with our memory-routed multi-head attention in the decoder block of our HybridNet. Unless otherwise noted, settings are the same for all experiments. During the training, we set the batch size to be 128 for one GPU and use the Adam [17] optimizer with 5000 warm-up steps, and learning rate initialized at 1e-4, and a dropout probability of 0.1 after the residual layer. It takes about 10 hours for the training set. The number of epoch is 800. We set the hyper-parameters for linear combination of attention maps in our experiments, including $\alpha$=0.1, $\beta$=0.4 and $\gamma$=0.1. During the test time, we validate learning outcomes after each learning epoch and select the model weights with the best CIDEr as our final results.

### 4.3 Results and Comparisons

**Quantitative Results.** We report our state-of-the-art results of the test on V2C [11] dataset for two tasks in Table 1. The previous works individually generate different types of commonsense captions by using separate networks. They can not predict all commonsense descriptions in a unified way. However, our HybridNet can generate all commonsense captions and benefit from their interaction.

On completion task, we used the reported results [11] for a fair comparison. As we can see, our HybridNet achieves the best performance on all metrics compared with the state-of-the-art methods [11, 12, 37, 45]. On the attribute part, our HybridNet performs 11.4% improvement on BLEU-1 better than the video CMS transformer [11]. On the effect and intention part, our network tends to bring better gains on the metrics that is applied for evaluating long and logical sentences, like the improvements on BLEU-4 (e.g., +19.0%, +18.4%) are better than the BLEU-1 (e.g., +8.6%, +14.2%). It can validate that our network can capture well the long-range context to make the semantics of the generated long sentences more reasonable and logical. The possible reason is our MMHA can learn the correlation between history sequence and next word prediction, which is cooperated with our memory module and contextual residual connections for word-level reasoning.

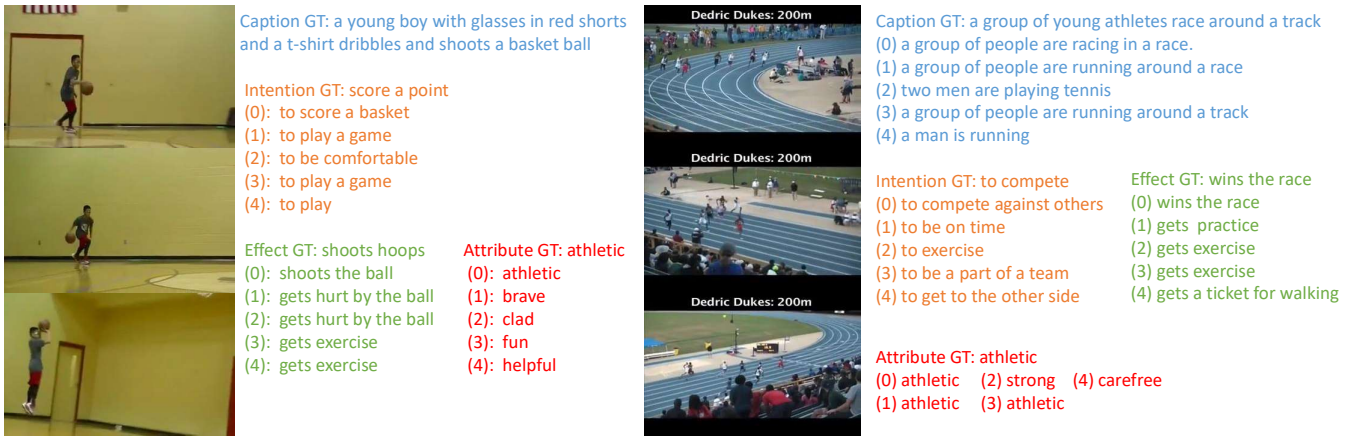

Figure 4: Qualitative visual results on completion task (left) and generation task (right). The (0)-(4) denote the prediction results of our HybridNet, Video CMS Transformer [11], Dense Captioner [45], S2VT [37] and Attention-Enc-Dec [12] respectively. Compared with the other state-of-the-art methods, our HybridNet can generate more precise and logical descriptions.

On the generation task, we re-implement and train the previous methods by using their official codes. Then we use some evaluation metrics that are widely used and accepted for language generation task to objectively estimate the performance of different methods.[2] In Table 1, our HybridNet achieves the best performances of 34.2%, 73.2%, 57.4%, 37.2%, 26.3% and 58.3% with respect to 7 metrics (i.e. CIDEr, BLEU-1 to BLEU-4, Meteor and Rouge-L) compared with 4 state-of-the-art methods on effect+C part. Our method also gets the best scores on other evaluation metrics. It can further prove that our network not only can perform well on the completion task but also can handle the more challenging task like the generation task.

**Qualitative Results.** Figure 4 shows the comparison results by different methods on completion task on the left and generation task on the right. On the completion task, our HybridNet can predict more precise intention results such as "to score a basket" compared with other methods. As we can see, other methods mainly focus on the vague intention expression (e.g., play a game) rather than the specific sports. On the generation task, our network still performs the best to jointly predict all the correct results. Although some methods can generate the correct caption (e.g., running around a race) and attribute (e.g., athletic), they still fail to provide the right intention and effect. On the contrary, thanks to the correct parsing into the caption and attribute, our method can also correctly provide reasonable and logical effect result (e.g., wins the race) and locate at the accurate intention (e.g., compete against others). Such semantic-level interaction can be achieved by our multi-commonsense learning, which can further demonstrate that our model can successfully learn the constraint and interaction of the multiple commonsense semantics.

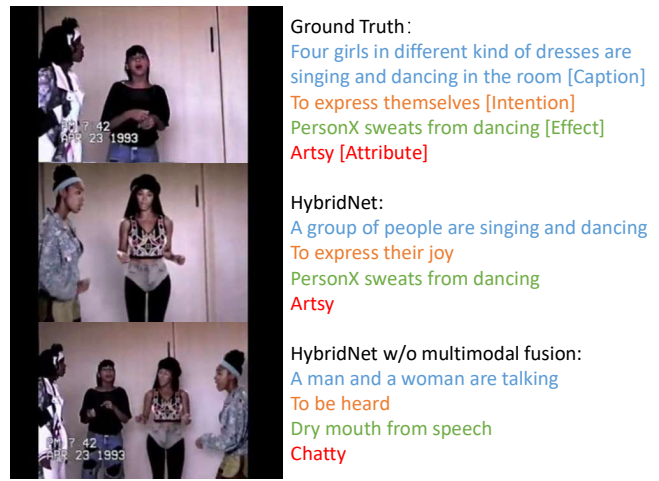

Figure 5: Visualization to compare the difference between HybridNet and HybridNet without multimodal fusion. The multimodal signal like audio in this example is necessary to distinguish between singing and talking.

### 4.4 Ablation Studies

**Effect of Multimodal Features.** In Table 2, the advantage of multimodal fusion increases BLEU-1 by 0.1% (58.6% vs 58.7%) on attribute completion, 0.7% on effect completion and 0.7% on intention completion. In Figure 5, we display the effectiveness and importance of multimodal fusion in the video-based commonsense captioning task. As we can see, the model guided with the audio feature and motion feature can better recognize the singing and dancing rather than talking. Based on the assistance of the multimodal fusion, our network can easily infer the correct commonsense descriptions.

---
[2]The previous methods only use the human evaluation for the generation task, which is too subjective.

Table 2: Ablation studies on two sub-tasks. The "multimodal", "multi-cms", "MMHA" and "CRC" means the multimodal fusion, multi-commonsense learning, memory-routed multi-head attention and contextual residual connections, respectively.

| | Relation | multimodal | multi-cms | MMHA | CRC | CIDER | BLEU-1 | BLEU-2 | BLEU-3 | BLEU-4 | METEOR | ROUGE-L |
|---|---|---|---|---|---|---|---|---|---|---|---|---|
| **Completion Task** | Attribute | | | | | - | 47.3 | - | - | - | - | - |
| | | | ✓ | | | - | 56.5 | - | - | - | - | - |
| | | ✓ | ✓ | | | - | 57.6 | - | - | - | - | - |
| | | ✓ | ✓ | ✓ | | - | 57.9 | - | - | - | - | - |
| | | | ✓ | ✓ | ✓ | - | 58.6 | - | - | - | - | - |
| | | ✓ | ✓ | ✓ | ✓ | - | **58.7** | - | - | - | - | - |
| | Effect | | | | | 37.3 | 34.8 | 25.9 | 22.5 | 20.4 | 20.8 | 30.6 |
| | | | ✓ | | | 60.6 | 45.6 | 39.0 | 35.9 | 34.8 | 28.0 | 38.9 |
| | | ✓ | ✓ | | | 63.1 | 45.8 | 39.2 | 36.0 | 33.9 | 28.4 | 40.0 |
| | | ✓ | ✓ | ✓ | | 64.8 | 47.2 | 40.9 | 38.0 | 36.2 | 29.2 | 40.7 |
| | | | ✓ | ✓ | ✓ | 65.8 | 48.3 | 42.3 | 39.7 | 38.0 | 29.4 | 41.0 |
| | | ✓ | ✓ | ✓ | ✓ | **66.2** | **49.0** | **42.9** | **40.3** | **38.8** | **30.0** | **41.5** |
| | Intention | | | | | 62.0 | 60.8 | 48.4 | 39.1 | 34.1 | 28.5 | 54.6 |
| | | | ✓ | | | 84.0 | 66.7 | 57.7 | 51.5 | 49.1 | 33.9 | 58.1 |
| | | ✓ | ✓ | | | 86.4 | 67.3 | 57.7 | 51.9 | 49.2 | 34.4 | 58.6 |
| | | ✓ | ✓ | ✓ | | 91.8 | 68.7 | 59.3 | 53.7 | 51.1 | 35.5 | 60.1 |
| | | | ✓ | ✓ | ✓ | 92.0 | 68.7 | 59.5 | 53.8 | 51.1 | 35.6 | 60.1 |
| | | ✓ | ✓ | ✓ | ✓ | **92.6** | **69.4** | **60.5** | **55.4** | **53.1** | **35.8** | **60.1** |
| **Generation Task** | Attribute+C | | | | | 40.2 | 70.2 | 54.8 | 42.7 | 32.6 | 24.7 | 59.0 |
| | | | ✓ | | | 40.2 | 70.0 | 55.7 | 44.7 | 35.2 | 25.0 | 59.8 |
| | | ✓ | ✓ | | | 40.2 | 70.0 | 56.5 | 45.5 | 35.4 | 25.1 | 60.2 |
| | | ✓ | ✓ | ✓ | | 41.0 | 71.1 | 56.9 | 45.6 | 35.6 | 25.2 | 60.3 |
| | | | ✓ | ✓ | ✓ | 40.6 | 69.6 | 55.4 | 44.2 | 34.6 | 24.9 | 59.7 |
| | | ✓ | ✓ | ✓ | ✓ | **41.6** | **71.3** | **57.0** | **45.6** | **35.7** | **25.5** | **60.4** |
| | Effect+C | | | | | 32.1 | 72.5 | 56.1 | 44.3 | 35.2 | 25.6 | 57.4 |
| | | | ✓ | | | 32.3 | 72.4 | 56.7 | 45.8 | 36.0 | 25.9 | 57.7 |
| | | ✓ | ✓ | | | 32.4 | 72.8 | 57.0 | 46.1 | 36.4 | 26.1 | 58.1 |
| | | ✓ | ✓ | ✓ | | 33.1 | 72.9 | 57.2 | 46.2 | 36.8 | 26.3 | 58.2 |
| | | | ✓ | ✓ | ✓ | 32.7 | 71.9 | 56.4 | 45.6 | 37.1 | 26.0 | 57.8 |
| | | ✓ | ✓ | ✓ | ✓ | **34.2** | **73.2** | **57.4** | **46.3** | **37.2** | **26.3** | **58.3** |
| | Intention+C | | | | | 37.8 | 76.2 | 61.2 | 48.1 | 37.3 | 26.9 | 61.9 |
| | | | ✓ | | | 37.7 | 76.7 | 62.0 | 49.7 | 39.8 | 27.2 | 62.0 |
| | | ✓ | ✓ | | | 38.2 | 77.1 | 62.2 | 50.1 | 40.1 | 27.4 | 62.1 |
| | | ✓ | ✓ | ✓ | | 39.6 | 77.2 | 62.4 | 50.3 | 40.2 | 27.6 | 62.9 |
| | | | ✓ | ✓ | ✓ | 38.9 | 76.5 | 62.3 | 49.8 | 39.8 | 27.2 | 62.9 |
| | | ✓ | ✓ | ✓ | ✓ | **40.4** | **77.5** | **62.9** | **50.4** | **40.2** | **27.8** | **62.9** |

**Effect of Multi-Commonsense Learning.** The effect of our multi-commonsense learning is apparently to boost BLEU-1 by around 9.2% (47.3%→56.5%), 10.8% (34.8%→45.6%) and 5.9% (60.8%→66.7%) from the baseline on the completion task w.r.t. attribute, effect and intention in Table 2. The effectiveness of our multi-commonsense learning can also be generalized to the generation task. It gets great scores on BLEU-3 and BLEU-4 compared with the baseline (e.g., 42.7%→44.7%, 44.3%→45.8% and 48.1%→49.7% on the attribute, effect and intention in terms of BLEU-3). These significant improvements can support the advantage and effectiveness of our multi-commonsense learning for semantic-level reasoning.

**Effect of Memory-routed Multi-Head Attention (MMHA).** In Table 2, we observe that our MMHA can bring improvements on all metrics in terms of the completion and generation tasks. For example, our MMHA can promote the network on the intention generation from 38.2% to 39.6% on CIDEr, and from 62.1% to 62.9% on Rouge-L. And it can reach a higher improvement of 5.4% CIDEr on the intention completion. In Figure 6 (a), the model with the MMHA can successfully reason the correlation between the pronoun "it" and the noun "foundation" since it remembers the history sequence that is applied to infer the next word. However, in Figure 6 (b), the model without MMHA is failed to reason the word-level correlation between "it" and "foundation". Besides, when predicting the pronoun "her", our MMHA infers the logical connection with the "woman", while the model without MMHA ignores the history memory and wrongly connects "her" with "a". The quantitative and qualitative results demonstrate the effectiveness and availability of our MMHA in video-based commonsense reasoning.

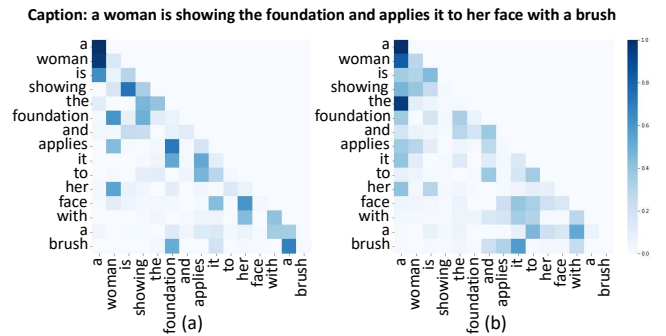

Figure 6: (a) Visualization of the attention map from our MMHA. (b) Visualization of the attention map from the traditional multi-head attention.

**Effect of Contextual Residual Connection.** Our contextual residual connection (CRC) aims to enhance the attention maps with a global perspective between different layers. In Table 2, our CRC collaborated with other modules is evaluated on two tasks and increases 0.8% CIDEr (39.6%→40.4%) and 2.6% BLEU-4 (36.2%→38.8%) on intention generation and effect completion, respectively. It can support the feasibility of incorporating the contextual residual connection into the decoder of our HybridNet.

## 5 CONCLUSION

In this paper, we propose a novel Hybrid Reasoning Network (HybridNet) for video-based commonsense captioning, which is jointly optimized by semantic-level reasoning and word-level reasoning. Multi-commonsense learning is built to achieve the semantic-level reasoning by jointly training different commonsense types in a unified network. To perform the word-level reasoning, a memory-routed multi-head attention (MMHA) is presented to inject the previous relational states from the memory module into the decoder for updating the attention maps. Then the updated attention maps are merged with other attention maps from previous decoder blocks by using contextual residual connections. The final obtained attention maps can be used to evolve representation. Our HybridNet achieves a new state-of-the-art on the large-scale Video-to-Commonsense benchmark and provided abundant analysis to demonstrate the effectiveness of our proposed modules.

## 6 ACKNOWLEDGEMENTS

This work was supported in part by National Natural Science Foundation of China under Grant No.U1811461, in part by the National Key Research and Development Project under Grant No.2020AAA0106600, in part by Natural Science Foundation of Guangdong Province, China under Grant No.2018B030312002, and in part by the Major Program of Guangdong Basic and Applied Research No.2019B030302002.

Table 3: Ablation studies on different fusion for multimodal information on completion task.

|  | MLP Fusion | Concat Fusion | CIDER | BLEU-1 | BLEU-2 | BLEU-3 | BLEU-4 | METEOR | ROUGE-L |
|---|---|---|---|---|---|---|---|---|---|
| Attribute | ✓ |  | - | 58.1 | - | - | - | - | - |
|  |  | ✓ | - | 58.7 | - | - | - | - | - |
| Effect | ✓ |  | 65.8 | 47.6 | 41.9 | 38.7 | 37.2 | 29.5 | 40.6 |
|  |  | ✓ | 66.2 | 49.0 | 42.9 | 40.3 | 38.8 | 30.0 | 41.5 |
| Intention | ✓ |  | 91.8 | 68.9 | 59.8 | 54.7 | 51.9 | 35.6 | 60.1 |
|  |  | ✓ | 92.6 | 69.4 | 60.5 | 55.4 | 53.1 | 35.8 | 60.1 |

Table 4: Quantitative results of the example in figure 5

|  | Multimodal | CIDER | BLEU-1 | BLEU-2 | BLEU-3 | BLEU-4 | METEOR | ROUGE-L |
|---|---|---|---|---|---|---|---|---|
| Attribute | ✓ | 40.7 | 69.2 | 56.9 | 44.9 | 35.2 | 25.2 | 58.4 |
|  |  | 39.5 | 68.3 | 55.6 | 43.8 | 34.6 | 24.8 | 57.3 |
| Effect | ✓ | 33.8 | 72.1 | 56.2 | 45.8 | 36.7 | 25.4 | 57.9 |
|  |  | 32.6 | 72.0 | 55.8 | 44.7 | 35.9 | 24.6 | 56.8 |
| Intention | ✓ | 40.2 | 76.8 | 61.5 | 49.8 | 40.1 | 27.9 | 61.6 |
|  |  | 39.8 | 76.2 | 61.3 | 49.4 | 39.5 | 27.2 | 61.2 |

## A APPENDIX

### A.1 Speed and Parameter Comparison

We have tested the inference time of [11] and our models by using the FLOPs metric. The CMS Transformer [11] needs 4.55 GFLOPs and our model only needs 2.93 GFLOPs. It can demonstrate the [11] is more time-consuming compared with our model. Besides, the parameters of our model are 103.4M, which is smaller than CMS (159.1M).

### A.2 Quantitative Analysis of the example in Figure 5

We tested the quantitative results of the example in Figure 5 between our model and [11]. The results are reported in Table 4. The example comes from the generation task. On attribute part, the [11] achieves 68.3 BLUE-1 and our model performs 69.2. Moreover, our model outperforms [11] by 1.1% Rough-L (57.9 vs 56.8) on effect part, and by 0.7% Meteor on intention part.

### A.3 The Effect of Different Fusion for Multimodal Inputs

For the encoder part, we mainly focus on the multimodal inputs. Hence the three pretrained encoders aim to extract multimodal features (e.g., 1D, 2D, 3D). In addition, we have done experiments to compare our fusion method in the encoder part with other fusion way. We replace our concatenation with MLP layers and train the whole network on the completion task. As shown in Table 3, the MLP fusion method achieves 65.8 Cider on effect and 91.8 on intention. Our concatenation performs 66.2 Cider on effect and 92.6 on intention. We believe that simply stacking parameters to increase the fusion block (MLP fusion) will not necessarily improve the video captioning. In other words, the multimodal inputs are the key to improve the final performance.

### A.4 Human Evaluation

We follow the human evaluation setting in [11] and hire 5 students for estimate the results for each model. We estimate our model on generation task with human evaluations and also compare it with the gold annotations from [11]. Our proposed model gets 66.45/78.96/71.78 on Effect/Attribute/Intention by using human evaluation. The Gold Annotations of human evaluation in [11] are 75.19/83.03/80.11 based on the ground truth.